\documentclass[runningheads]{llncs}

\usepackage[width=122mm,left=12mm,paperwidth=146mm,height=193mm,top=12mm,paperheight=217mm]{geometry} 

\usepackage{graphicx}
\usepackage{multirow}
\usepackage{amsmath}
\usepackage{amssymb}
\usepackage{bbm}
\usepackage{array,multirow}
\usepackage{arydshln}
\usepackage{makecell}
\usepackage{siunitx}
\usepackage{etoolbox}        
\usepackage{xcolor}
\newrobustcmd{\R}{\color{red}}
\newrobustcmd{\B}{\bfseries}
\newrobustcmd{\T}{\scriptsize}

\newcommand{\beginsupplement}{%
        \setcounter{table}{0}
        \renewcommand{\thetable}{S\arabic{table}}%
        \setcounter{figure}{0}
        \renewcommand{\thefigure}{S\arabic{figure}}%
     }

\begin{document}
\title{Unsupervised Domain Adaptation with Contrastive Learning for OCT Segmentation}
\titlerunning{Unsupervised Domain Adaptation with Contrastive Learning}

\author{Alvaro Gomariz\inst{1} \and
Huanxiang~Lu\inst{1} \and
Yun~Yvonna~Li\inst{1} \and
Thomas~Albrecht\inst{1} \and
Andreas~Maunz\inst{1} \and
Fethallah~Benmansour\inst{1} \and
Alessandra~M.~Valcarcel\inst{2} \and
Jennifer~Luu\inst{2} \and
Daniela~Ferrara\inst{2} \and
Orcun~Goksel\inst{3,4}}

\authorrunning{Gomariz et al.}

\institute{
F Hoffmann-La Roche AG, Basel, Switzerland
\and
Genentech Inc, California, United States
\and
Computer-assisted Applications in Medicine, ETH Zurich, Zurich, Switzerland \and
Department of Information Technology, Uppsala University, Uppsala, Sweden
}

\maketitle   
\begin{abstract}
Accurate segmentation of retinal fluids in 3D Optical Coherence Tomography images is key for diagnosis and personalized treatment of eye diseases. 
While deep learning has been successful at this task, trained supervised models often fail for images that do not resemble labeled examples, e.g.\ for images acquired using different devices. 
We hereby propose a novel semi-supervised learning framework for segmentation of volumetric images from new unlabeled domains.
We jointly use supervised and contrastive learning, also introducing a contrastive pairing scheme that leverages similarity between nearby slices in 3D.
In addition, we propose channel-wise aggregation as an alternative to conventional spatial-pooling aggregation for contrastive feature map projection.
We evaluate our methods for domain adaptation from a (labeled) source domain to an (unlabeled) target domain, each containing images acquired with different acquisition devices.
In the target domain, our method achieves a Dice coefficient 13.8\% higher than SimCLR (a state-of-the-art contrastive framework), and leads to results comparable to an upper bound with supervised training in that domain. 
In the source domain, our model also improves the results by 5.4\% Dice, by successfully leveraging information from many unlabeled images.
\end{abstract}

\section{Introduction}
% Motivation: ophthalmology/OCT/need for segmentation
Supervised learning methods, in particular UNet~\cite{unet}, for segmentation of retinal fluids imaged with Optical Coherence Tomography (OCT) devices have led to major advances in diagnosis, prognosis, and understanding of eye diseases~\cite{bogunovic2019retouch,unetretina18,fujimoto2016development,sahni2019machine,schmidt2016paradigm}. 
However, training these supervised deep neural networks requires large amounts of labeled data, which are costly, not always feasible, and need to be repeated for each problem domain; since
trained models often fail when inference data differs from labeled examples, so-called \emph{domain-shift}, e.g.\ for images from a different OCT device~\cite{schlegl2018fully}. 
%
% Domain adaptation
Unsupervised domain adaptation aims to leverage information learned from a labeled data domain for applications in other domains where only unlabeled data is available. 
To this end, many deep learning methods have been proposed~\cite{wang2018deep}, mostly using generative adversarial networks, e.g.\ to translate visual appearance across OCT devices~\cite{ren2021segmentation}.

% Contrastive learning
Contrastive learning (CL) aims to extract informative features in a self-supervised manner by comparing (unlabeled) data pairs in a feature subspace of a network~\cite{caron2021emerging,simclr,chen2020big,chen2021exploring,grill2020bootstrap,he2020momentum,khosla2020supervised,oord2018representation}. 
A widely-adopted CL framework, SimCLR~\cite{simclr}, generates positive image pairs from the same image via image augmentations to minimize feature distances between these pairs, while maximizing their distance from augmentations of other images as negative samples.
Other CL strategies aim to successfully learn without a need for negative pairs, \mbox{SimSiam}~\cite{chen2021exploring} being a representative example. 
CL is commonly used for pretraining models, typically using natural images such as ImageNet~\cite{deng2009imagenet}, which are then finetuned or distilled for downstream tasks, e.g.\ classification, detection, or segmentation~\cite{chen2020big}.

% Medical images
Models pretrained with natural images are of limited use for medical applications, which involve images with substantially different appearances and often with 3D content, leading to a recent focus on application-specific approaches for CL pair generation in medical context~\cite{chaitanya2020contrastive,chen2021uscl}.
USCL~\cite{chen2021uscl} minimizes the feature distance between frames of the same ultrasound video, while maximizing the distance between frames of different videos, in order to produce pretrained models for ultrasound applications. 
USCL also proposes a joint semi-supervised approach, which simultaneously minimizes a contrastive and supervised \emph{classification} loss.
However, to be applicable for image segmentation, this method relies on subsequent finetuning, which is potentially sub-optimal for preserving the unlabeled information for the intended task of segmentation.
In fact, there exist little work on CL methods on image segmentation without finetuning.

% Contributions
We hereby aim to improve segmentation quality of OCT datasets with limited manual annotations, but with abundant unlabeled data.
We focus on unsupervised domain adaptation, where manual annotations exist for one device (source domain), but not for another (target domain).
We achieve this with the following contributions: $\bullet$ We introduce a semi-supervised framework for joint training of CL together with segmentation labels
(Section~\ref{sec:framework}). $\bullet$ We propose an augmentation strategy that leverages expected similarity between nearby slices in 3D (Section~\ref{sec:augmentation}). $\bullet$ We introduce a new CL projection head (Section~\ref{sec:projection}) that aggregates features without losing spatial context, which produces results superior to the conventional spatial pooling strategy. 
Our contributions are tested on two large clinical datasets collected in trials using different OCT imaging devices.

\section{Methods}
\subsection{Simultaneously learning from labeled and unlabeled data}
\label{sec:framework}
% Supervised learning
As the segmentation backbone, we utilize the proven UNet architecture~\cite{unet}, which can be modeled as $F(\cdot)$ processing an image $x$ to produce a segmentation map $p=F(x)$ to approximate an (expert-annotated) ground truth segmentation~$y$. 
In the supervised setting, $F$ is learned by minimizing a supervised loss $\mathcal{L}_\mathrm{sup}$, which is for us the logarithmic Dice loss of labeled data in a domain $D$:
\begin{equation}
    \mathcal{L}_\mathrm{sup} = -\sum_{(p_i,y_i) \in D}\log{\frac{2 \sum_{j \in \mathrm{pixels}} y_i^j p_i^j}{\epsilon + \sum_{j \in \mathrm{pixels}}  (y_i^j + p_i^j)}}
\end{equation}
for all training images $i$ in $D$, where $\epsilon$ is a small number to avoid division by 0.

\begin{figure}[t]
\includegraphics[width=\textwidth]{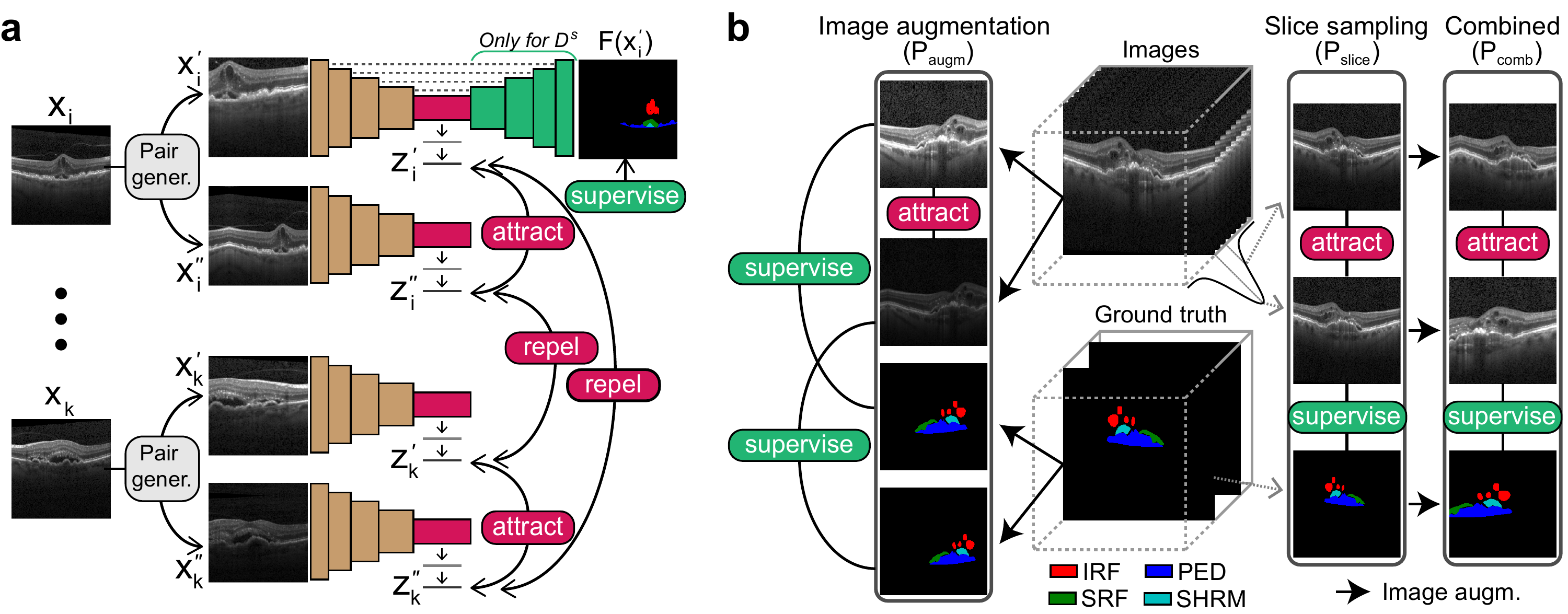}
\caption{Illustration of our CL methods.
(a) Semi-supervised contrastive learning framework for unsupervised domain adaptation. 
Note that the \emph{repel} modules do not apply to SimSiam. 
(b) Proposed pair generation methods for contrastive learning on 3D images.
}
\label{fig:network_augmentation}
\end{figure}

% Contrastive learning
Contrastive frameworks aim to learn features $h=E(x)$ with an encoder $E(\cdot)$ without the need of manually annotated labels $y$. 
We herein base our methods on the SimCLR framework~\cite{simclr}. 
In order to adapt the learned features $h$ for our intended segmentation task, we replace the originally-proposed ResNet architecture for $E(\cdot)$ with the UNet encoder (illustrated in brown in Fig.\,\ref{fig:network_augmentation}a).
A subsequent contrastive projection head $C(\cdot)$ maps the bottleneck-layer features to vector projections $z=C(h)$ on which the contrastive loss $\mathcal{L}_\mathrm{con}$ is applied. 
This loss aims to minimize the distance between ``positive'' pairs of images $(x_i',x_i'')$ created from each image $x_i$ by a defined pair generator $P(\cdot)$ described further in Section~\ref{sec:augmentation} below, i.e.\ $P(x_i)=(x_i',x_i'')$. 
We employ a version of the normalized temperature-scaled cross entropy loss~\cite{oord2018representation} adapted to our problem setting as:
\begin{gather}
    L_\mathrm{con}^\mathrm{CLR}= \sum_{P(x_i),\ x_i \in D}\big(\,l(z_i', z_i'') + l(z_i'', z_i')\,\big)\\
    l(z_i',z_i'') = -\log \frac{\exp{\big(d(z_i',z_i'')}/\tau\big)}{\sum_{x_i \in D} \mathbbm{1} _{[k \neq i]}\exp{\big(d(z_i',z_k'')/\tau\big)}}
\end{gather}
where $d(u,v) = (u \cdot v) /  (||u||_2\, ||v||_2)$ and $\tau$ is the temperature scaling parameter. 

In SimSiam, a learnable predictor $Q(\cdot)$ is applied on one projection to predict the other:
\begin{equation}
     L_\mathrm{con}^\mathrm{Siam} = -\sum_{x_i \in D}\Big(d\big(Q(z_i'), z_i''\big) + d\big(Q(z_i''), z_i'\big)\Big)
\end{equation}
where the gradients from the second projection pairs are prevented from back-propagating for network weight updates (\emph{stopgrad}).

We adapt the USCL joint training strategy, which was proposed for US video classification, to our segmentation task on 3D images by combining $\mathcal{L}_\mathrm{sup}$ and $\mathcal{L}_\mathrm{con}$ in a semi-supervised framework illustrated in Fig.~\ref{fig:network_augmentation}a.
Considering a source domain $D^\mathrm{s}$ and a target domain $D^\mathrm{t}$, total loss $\mathcal{L}$ is calculated as follows:
\begin{equation}
    \mathcal{L} = \frac{1}{2} \left( \underset{x \in D^\mathrm{s}}{\mathcal{L}_\mathrm{con}} + \underset{x \in D^\mathrm{t}}{\mathcal{L}_\mathrm{con}} \right) + \lambda \underset{(x,y) \in D^\mathrm{s}}{\mathcal{L}_\mathrm{sup}}
\end{equation}

\subsection{Pair generation strategy}
\label{sec:augmentation}
Generation of pairs for the contrastive loss is key for successful self-supervised learning. 
We herein propose and compare different pair generation functions $P(\cdot)$ for volumetric OCT images, as illustrated in Fig.~\ref{fig:network_augmentation}b.

We denote by $P_\mathrm{augm}$ an OCT adaptation of the pair formation typically employed for natural images (e.g., in SimCLR and SimSiam). 
Here, labeled slices in $D^\mathrm{s}$ and random slices in $D^\mathrm{t}$ are augmented with horizontal flipping ($p=0.5$), horizontal and vertical translation (within 25\% of the image size), zoom in (up to 50\%), and color distortion (brightness up to 60\% and jittering up to 20\%). 
For color augmentation, images are transformed to RGB, and then back to grayscale.

We propose $P_\mathrm{slice}$ that leverages the coherence of nearby slices in a 3D volume for CL.
Here, $x_i'=x_i$ for a slice index $b_i'$ in 3D. Then, $x_i''$ is a slice from the same volume with the (rounded) slide index $b_i'' \sim \phi(b_i',\sigma)$, where $\phi$ is a Gaussian distribution centered on $b_i'$, with standard deviation $\sigma$ as a hyperparameter. 
Combining the two pairing strategies yields $P_\mathrm{comb}$ where $P_\mathrm{slice}$ is used first and the augmentations in $P_\mathrm{augm}$ are then applied on the selected slices.

\subsection{Projection heads to extract features for image segmentation}
\label{sec:projection}
A projection head $C(\cdot)$ is formed by an aggregation function $\rho^\mathrm{agg}$ that aggregates features $h$ to form a vector, which is then processed by a multilayer perceptron $\rho^\mathrm{MLP}$ to create projection $z$. 
Typical contrastive learning frameworks, e.g.\ SimCLR and SimSiam, use a projection (denoted herein by $C_\mathrm{pool}$) where $\rho_\mathrm{pool}^\mathrm{agg}:  \mathbb{R}^{w \times h \times c} \rightarrow \mathbb{R}^{1 \times 1 \times c}$ is a global pooling operation on the width $w$, height $h$, and channels $c$ of the input features.
Such projection $C_\mathrm{pool}$ may be suboptimal for learning representations to effectively leverage segmentation information, as backpropagation from $\mathcal{L}_\mathrm{con}$ would lose the spatial context. 
Instead we propose $C_\mathrm{ch}$, for which $\rho_\mathrm{ch}^\mathrm{agg}: \mathbb{R}^{w \times h \times c} \rightarrow \mathbb{R}^{w \times h \times 1}$ is a $1$$\times$$1$$\times$$1$ convolutional layer that learns how to aggregate layers, so the spatial context is preserved. 

\section{Experiments and Results}
\noindent{\bf Dataset. }
We employ two large OCT datasets from clinical trials on patients with neovascular age-related macular degeneration.
Images acquired using a \emph{Spectralis} (Heidelberg Engineering) imaging device have $512 \times 496 \times 49$ or $768 \times 496 \times 19$ voxels, with a resolution of $10 \times 4 \times 111$ or $5 \times 4 \times 221$ $\mu$m/voxel, respectively.
These were acquired as part of the phase-2 AVENUE trial (NCT\,02484690). 
Images acquired as part of another study, phase-3 HARBOR trial (NCT\,00891735), were acquired with a \emph{Cirrus} HD-OCT III (Carl Zeiss Meditec) imaging device, which produces scans with $512 \times 128 \times 1024$ voxels and a resolution of $11.7 \times 47.2 \times 2.0$ $\mu$m/voxel.
All slices (B-scans) from the two different devices are resampled to $512 \times 512$ pixels with roughly the same resolution of $10 \times 4$ $\mu$m/pixel.
Select B-scans from Spectralis were manually annotated for fluid regions of potential diagnostic value: intraretinal fluid (IRF), subretinal fluid (SRF), pigment epithelial detachment (PED), and subretinal hyperreflective material (SHRM). 
More details on these datasets and the annotation protocol can be found in~\cite{maunz21}.
In our experiments, we use all training data from Spectralis as source domain $D^\mathrm{s}$, and unlabeled images from Cirrus as target domain $D^\mathrm{t}$.
Labeled data from Cirrus is only used for the training of an \emph{UpperBound} model for $D^\mathrm{t}$.
Data stratification used in our evaluations is detailed in the supplementary Table~S1.

\vspace{1ex}\noindent{\bf Implementation. }
Adam optimizer~\cite{kingma2014adam} was used in all models, with a learning rate of $10^{-3}$.
Dropout with \mbox{$p=0.5$} is applied before and after each convolutional block in the lowest UNet resolution level, as well as after the convolutions in the two subsequent resolution levels of the decoder. 
Group normalization~\cite{wu2018group} with 4 groups is used after each convolutional layer.  
After the aggregation function $\phi$ in $C(\cdot)$, two fully-connected layers are used with 128 units each, where the first one uses ReLU activation.
We heuristically set $\lambda=20$ and the standard deviation of $\phi$ for $P_\mathrm{slice}$ as \mbox{$\sigma=0.25\,\mu m$}, which is the range for which we observe roughly similar features across slices.
Implementation is in Tensorflow 2.7, ran on an NVIDIA V100 GPU. 

\vspace{1ex}\noindent{\bf Metrics. }
We segment individual slices with 2D UNet, since (1) only some slices were annotated in OCT volumes; and (2) this enables our slice-contrasting scheme.
Model performance was evaluated also slice-based, using the Dice coefficient and Unnormalized Volume Dissimilarity (UVD) on 2D slices.
The latter measures the extent of total segmentation error (FP+FN) in each slice and is more robust to FP on  B-scans with small annotated regions for individual classes.
Averaging metrics across classes with a large variation may lead to bias. Thus, we first normalize each per-slice metric ($m^c_i$) for method $i$ and class $c$ by its class Baseline ($m^c_\textrm{bas}$), and then average these over all $c$ and images on the test set.
All models with supervision were trained for 200 epochs, and
the model at the epoch with the highest average Dice coefficient across classes on the validation set was selected for evaluation on a holdout test set. 

\subsection{Evaluation on the unlabeled target domain}
We first evaluate our proposed methods in the desired setting of unsupervised domain adaptation; i.e.\ models trained on $(x,y) \in D^\mathrm{s}$ and $x \in D^\mathrm{t}$ are evaluated on $y \in D^\mathrm{t}$.
Note that, although unlabeled for training, $D^\mathrm{t}$ has some ground truth annotations in the test set to enable its evaluation (see Table~S1).
In Table~\ref{tab:results_models_rel} and Table~S2, \emph{UpperBound} results for a supervised model trained on labeled data from the target domain are also reported for comparison.
This labeled data, used here as a reference, is ablated for all other models. 
A supervised UNet model, \emph{Baseline}, was trained only on the source domain $D^\mathrm{s}$. Its poor performance on $D^\mathrm{t}$ confirms that the two domains indeed differ from supervised learning perspective.

\begin{table}[tb]
\centering
\caption{Evaluation on target domain $D^\mathrm{t}$ and source domain $D^\mathrm{s}$ across all classes, relative to Baseline (rel) and absolute values (abs), in red when metrics are inferior, and in bold for the best performance (excluding UpperBound).
Supervised methods use labels from the domain in brackets.
Dice is shown as \%, and UVD as $\mu m^3$x$10^{2}$.
}
\resizebox{\textwidth}{!}{%
\addtolength{\tabcolsep}{2pt}%
\sisetup{detect-all=true}%
\begin{tabular}{|@{\,}c@{\,}:l|@{}
S[table-format=-2.2]@{}
>{\scriptsize}S[table-format=-2.2]:
S[table-format=-2.2]@{}
>{\scriptsize}S[table-format=-2.2]|@{}
S[table-format=-2.2]@{}
>{\scriptsize}S[table-format=-2.2]:@{}
S[table-format=-2.2]@{}
>{\scriptsize}S[table-format=-2.2]|
}%
\hline
\multirow{2}{*}{Approach} & \multicolumn{1}{c|}{\multirow{2}{*}{ Methods }} & \multicolumn{4}{c|}{Domain $D^\mathrm{t}$} & \multicolumn{4}{c|}{Domain $D^\mathrm{s}$} \\
\cdashline{3-10}
& & \multicolumn{2}{c:}{ Dice rel ({\T abs}) } & \multicolumn{2}{c|}{ UVD rel ({\T abs}) } & \multicolumn{2}{c:}{ Dice rel ({\T abs}) } & \multicolumn{2}{c|}{ UVD rel ({\T abs}) } \\
\hline
\multirow{2}{*}{Supervised} & UpperBound[$D^\mathrm{t}$] & 29.32 & 63.88 & -8.93 & 8.67 & {-} & {-} & {-} & {-} \\
& Baseline[$D^\mathrm{s}$] & 0.00 & 34.57 & 0.00 & 17.60 & 0.00 & 67.36 & 0.00 & 5.80 \\
\hdashline
\multirow{2}{*}{Adversarial} & CycleGAN~\cite{seebock2019using} & \R -6.53 & \R 28.04 & \R 2.51 & \R 20.10 & \R -35.13 & \R 32.23 & \R 7.62 & \R 13.42 \\
& DAN~\cite{Bolte_2019_CVPR_Workshops} & 17.93 & 52.49 & -5.25 & 12.34 & \R -0.51 & \R 66.85 & \R 0.02 & \R 5.82 \\
\hdashline
\multirow{2}{*}{\makecell{Finetuning\\ \T (CL $\rightarrow$ supervision)}} & SimCLR~\cite{simclr} & 14.01 & 48.58 & -4.24 & 13.36 & \R -3.48 & \R 63.88 & \R 0.48 & \R 6.28 \\
& SimSiam~\cite{chen2021exploring} & 11.41 & 45.97 & -2.39 & 15.21 & 0.40 & 67.75 & \R 0.19 & \R 6.00 \\
\hdashline
\multirow{5}{*}{\makecell{Joint\\ \T (CL + supervision)}} & SegCLR($P_\mathrm{augm}$,$C_\mathrm{pool}$) & 23.22 & 57.78 & -5.91 & 11.68 & \R -0.65 & \R 66.71 & 0.00 & 5.80 \\
& SegSiam($P_\mathrm{augm}$,$C_\mathrm{pool}$) & \R -21.90 & \R 12.67 & \R 48.09 & \R 65.69 & \R -46.58 & \R 20.78 & \R 48.31 & \R 54.11 \\
& SegCLR($P_\mathrm{slice}$,$C_\mathrm{pool}$) & 6.14 & 40.71 & -2.81 & 14.79 & \R -15.14 & \R 52.22 & \R 2.26 & \R 8.06 \\
& SegCLR($P_\mathrm{comb}$,$C_\mathrm{pool}$) & 27.21 & 61.77 & -6.25 & 11.34 & 1.48 & 68.83 & \R 0.18 & \R 5.98 \\
& SegCLR($P_\mathrm{comb}$,$C_\mathrm{ch}$) & \B 27.77 & \B 62.33 & \B -6.71 & \B 10.88 & \B 1.93 & \B 69.28 & \B -0.09 & \B 5.71 \\
\hline
\end{tabular}%
}
\label{tab:results_models_rel}
\end{table}

\noindent{\bf Adversarial} approaches are included as state-of-the-art baselines for unsupervised domain adaptation. 
CycleGAN~\cite{seebock2019using} is adapted to our UNet using entire slices. 
Training converged with meaningful translated images from $D^\mathrm{t}$ to $D^\mathrm{s}$, on which we run the pretrained UNet.
Domain Adversarial Neural Network (DANN) includes a gradient reversal layer~\cite{ganin2015unsupervised} with the design in~\cite{Bolte_2019_CVPR_Workshops} for segmentation. 
While DANN performs better than Baseline on $D^t$,
CycleGAN is inferior. Our latter observation is contrary to that reported in~\cite{seebock2019using}, which is likely due to our Baseline being much superior to that of~\cite{seebock2019using} (with a reported Dice of near zero). 

\noindent{\bf Finetuning.} Learning representations of $D^\mathrm{t}$ with SimCLR and SimSiam with subsequent finetuning on $D^\mathrm{s}$ shows a clear improvement over Baseline for all classes, confirming that these CL strategies are also valid when adapted to our OCT dataset.
SimCLR produces better results than SimSiam, suggesting that the use of negative pairs helps in learning better representations in our case. 

\noindent{\bf Joint training} using the SimCLR framework and our above changes for a supervised loss for segmentation is herein called \emph{SegCLR} (\emph{SegSiam} for the SimSiam equivalent), which increases the number of parameters merely by 6.85\% (7.33\% for SegSiam). 
SegCLR($P_\mathrm{augm}$,$C_\mathrm{pool}$) shows an overall improvement over finetuning.
This is not the case for SegSiam($P_\mathrm{augm}$,$C_\mathrm{pool}$), which suggests that the lack of negative pairs makes it difficult to simultaneously optimize $\mathcal{L}_\mathrm{sup}$  and $\mathcal{L}_\mathrm{con}$; e.g.\ minimizing $\mathcal{L}_\mathrm{con}$ for only positive pairs may learn only simplistic features, which then would prevent $\mathcal{L}_\mathrm{sup}$ from improving features for segmentation. 

\noindent{\bf Pair generation. } $P_\mathrm{slice}$ alone produces poorer results compared to $P_\mathrm{augm}$ alone, indicating that merely contrasting nearby slices does not facilitate extracting features useful for segmentation. 
Nevertheless, by applying both pair generation methods together, i.e.\ with $P_\mathrm{comb}$, Dice and UVD results are overall superior to all the results above. 
This indicates that pairing nearby slices in our 3D images is a good complement to the typical image augmentation strategies.

\begin{figure}[t]
\centering
\includegraphics[width=\textwidth]{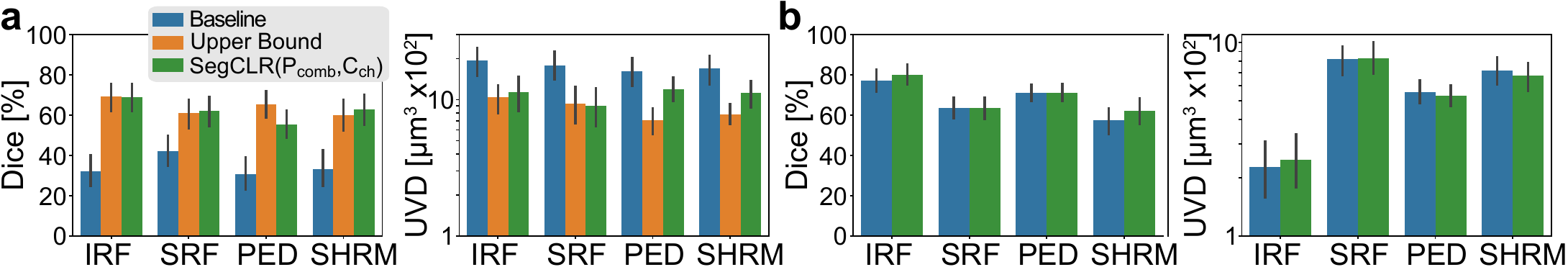}
\caption{
Evaluation of models for the different classes on (a) target domain $D^\mathrm{t}$ and (b)~source domain $D^\mathrm{s}$.
Black bars denote 95\% confidence intervals. 
} 
\label{fig:models_results_classes}
\end{figure}

\noindent{\bf Projections. } We change the typical $C_\mathrm{pool}$ head with our proposed $C_\mathrm{ch}$ designed specifically for the segmentation task, which adds a mere 0.03\% more parameters.
While for IRF and PED (Table S2) this performs worse than SegCLR($P_\mathrm{comb}$,$C_\mathrm{pool}$), the Dice and UVD metrics averaged across classes are overall the best for SegCLR($P_\mathrm{comb}$,$C_\mathrm{ch}$), notably even surpassing the UpperBound in some cases (Fig.\ \ref{fig:models_results_classes}a). 
Hence, our proposed model could replace the UpperBound if and when no training data is available in the target domain, and in doing so only compromising the performance for PED (Fig.\ \ref{fig:models_results_classes}a).

\subsection{Evaluation on the labeled source domain}
Herein we test the retention of segmentation information for the original source domain $D^\mathrm{s}$, as shown in Table~\ref{tab:results_models_rel} (right-most column) and Fig.\ \ref{fig:models_results_classes}b.
As expected, Baseline produces better results on $D^\mathrm{s}$ than on $D^\mathrm{t}$, since it is evaluated in the same domain in which it was supervised. 
For finetuning, contrary to its relative performance on $D^\mathrm{t}$, for $D^\mathrm{s}$ SimSiam  produces better results than SimCLR.
A reason could be SimSiam's use of only positive pairs leading to distinct features for each domain, which are later finetuned relatively more easily with segmentation supervision on $D^\mathrm{s}$. 
Further observations on $D^\mathrm{s}$ corroborate their above-discussed counterparts for $D^\mathrm{t}$; i.e.\ SegSiam fails; $P_\mathrm{slice}$ alone performs worse than $P_\mathrm{augm}$ alone; and combining them as $P_\mathrm{comb}$ performs the best. 
Our proposed SegCLR($P_\mathrm{comb}$,$C_\mathrm{ch}$) model produces the best results across classes also for this source domain $D^\mathrm{s}$, notably even surpassing the supervised Baseline. 
This shows that supervised information from the labeled domain is not forgotten (e.g.\ as a trade-off when learning from the unlabeled domain), but it is rather enhanced with the unlabaled data, despite the latter being from a different domain.

\begin{figure}[t]
\centering
\includegraphics[width=\textwidth]{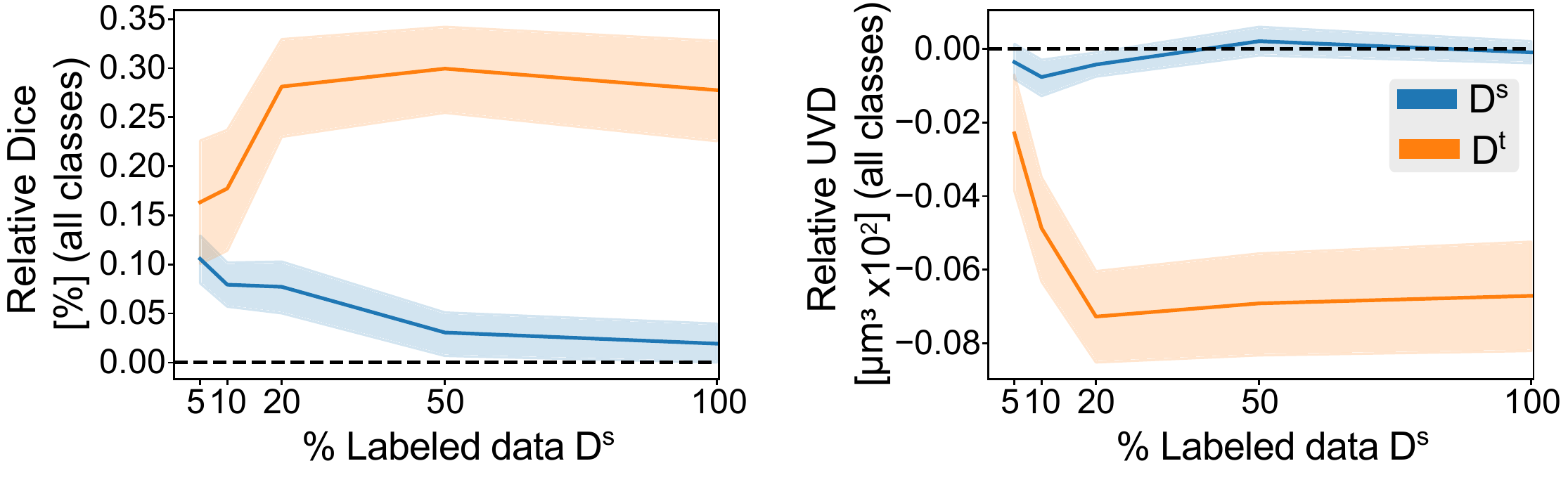}
\caption{Evaluation on $D^\mathrm{s}$ and  $D^\mathrm{t}$ datasets with models trained on 5, 10, 20, 50, and 100\% of labeled data from  $D^\mathrm{s}$.
Herein volume percentages are reported.
Results show the proposed SegCLR($P_\mathrm{comb}$,$C_\mathrm{ch}$) relative to Baseline with same \% of $D^\mathrm{s}$ labeled data. 
} 
\label{fig:data_ablation}
\end{figure}

\subsection{Ablations on amount of labeled data}
We study below the effect that the amount of labeled data in $D^\mathrm{s}$ has on the performance of our semi-supervised learning framework. 
To this end, we randomly ablate parts of the training data in $D^\mathrm{s}$.
The validation set was fixed to avoid any bias on model selection. 
Results in Fig.~\ref{fig:data_ablation} indicate that adding more labeled data from $D^\mathrm{s}$ in the training of our model has overall a positive effect on its effectiveness for segmentation of the target domain $D^\mathrm{t}$. 
This is likely because $\mathcal{L}_\mathrm{con}$ can adapt segmentation features to the $D^\mathrm{t}$ space only when these features are learned robustly with more labeled data, based on which $\mathcal{L}_\mathrm{sup}$ can be minimized.
The trend is somewhat the opposite for $D^\mathrm{s}$:
For the low data regime, $\mathcal{L}_\mathrm{con}$ seems to help with feature extraction, even though the information comes from a different domain. 
However, as the amount of labeled data increases and $\mathcal{L}_\mathrm{sup}$ is exposed to enough data from the source domain, any contrastive information contribution from a different unlabeled domain becomes relatively insignificant. 

\subsection{Segmentation results compared to inter-grader variability}
Manual annotation of retinal fluids is challenging, leading to large variability in segmentation metrics even among human experts.
We herein compare our proposed SegCLR($P_\mathrm{comb}$,$C_\mathrm{ch}$) to inter-grader discrepancies.
We employ a set of 44 OCT volumes, each fully annotated independently by 4 different graders.
These annotations are drawn from the same target domain $D^\mathrm{t}$ but come from a different clinical study than the dataset used in training, so a direct comparison is not possible.
We evaluated segmentation metrics for graders by comparing them with one another. 
We deem our method within inter-grader variability when its metric for a class and image, with respect to any grader, is better than that of at least one human inter-grader metric (variation). 
Across images and classes, SegCLR($P_\mathrm{comb}$,$C_\mathrm{ch}$) performs within such inter-grader variability in 65.34\% and 48.30\% of cases based on Dice and UVD, respectively.

\section{Conclusions}
Unsupervised domain adaptation for segmentation has been typically approached as finetuning on features learned via self-supervision from classification tasks.
We propose herein a segmentation approach that is jointly supervised with existing data while being self-supervised with abundant unlabeled examples from a previously unseen domain. 
With our proposed slice-based pairing and channel-wise aggregation for contrastive projections, our model successfully adapts supervised labeled-domain info to an unlabeled domain, surpassing previous state-of-the-art adversarial methods and even approaching the performance of an upper bound.
We also improve the results in the original labeled domain by leveraging the unsupervised (contrastive) info.
These contributions will help reduce manual annotation efforts for segmentation of 3D volumes in new data domains. 

\bibliographystyle{splncs04}
\bibliography{bibliography}

\newpage

\beginsupplement
\begin{center}
\Large {\bf{Supplementary Material\\}}
\end{center}
\begin{figure}[htb]
\centering
\includegraphics[width=0.85\textwidth]{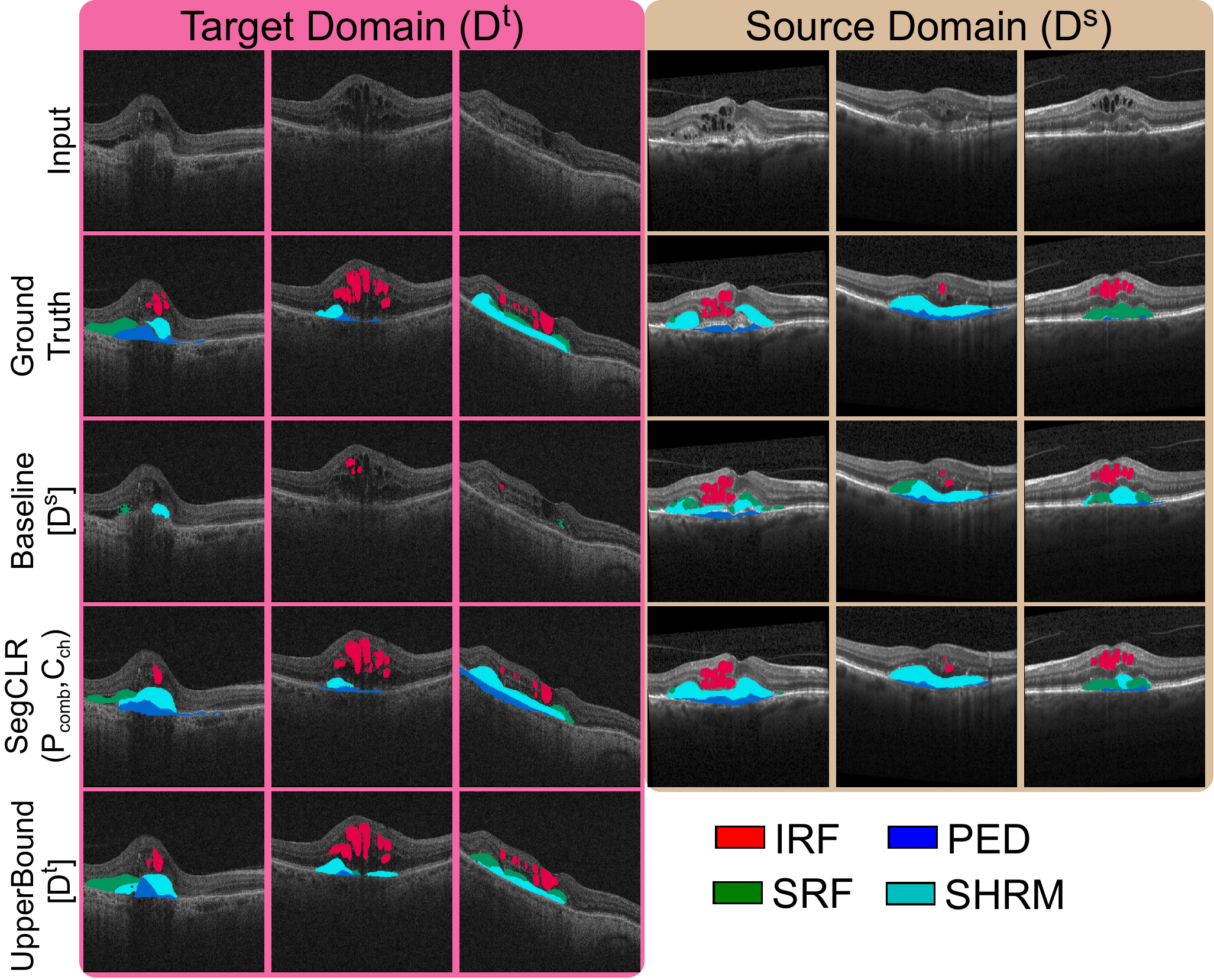}
\caption{
Qualitative assessment of segmentation on $D^\mathrm{t}$ and $D^\mathrm{s}$ examples.
}
\label{fig:visuals}
\end{figure}

\begin{table}[htb]
    \centering
    \caption{
    Datasets employed for the training and evaluation of models. 
    Labeled data for training is displayed as \#training+\#validation.
    Volumes from the Spectralis device were used as both labeled and unlabeled data, i.e.\ the annotated B-scans were used for $\mathcal{L}_\mathrm{sup}$, while all slices were available as unlabeled data for $\mathcal{L}_\mathrm{con}$.
    Labeled training data for Cirrus (denoted in parantheses) is used only for training UpperBound.
    }
    \scriptsize
\setlength{\extrarowheight}{.3ex}
\begin{tabular}{|c|c|cc|cc|cc|}
\hline
\multirow{3}{*}{ \textbf{Domain} } & \multirow{3}{*}{ \textbf{Device} } & \multicolumn{4}{c|}{\textbf{Training}} &
\multicolumn{2}{c|}{\textbf{Testing}} \\
\cdashline{3-8}
& & \multicolumn{2}{c|}{\textbf{Labeled}} & \multicolumn{2}{c|}{\textbf{Unlabeled}} & \multicolumn{2}{c|}{\textbf{Labeled}} \\
\cdashline{3-8}
& &\textbf{ B-scans} & \textbf{Volumes} & \textbf{B-scans} & \textbf{Volumes} & \textbf{B-scans} & \textbf{Volumes} \\
\hline
$D^\mathrm{s}$ & Spectralis & 1363+243 & 234+41 & 11\,466 & 275 & 163 & 28 \\
$D^\mathrm{t}$ & Cirrus & (735+125) & (122+21) & 6.8 million & 53\,197 & 99 & 17 \\
\hline
    \end{tabular}
    \label{tab:dataset}
\end{table}

\begin{table}[htb]
\caption{
Evaluation of models on target domain $D^\mathrm{t}$ for all 4 annotated classes.
This table corresponds to the results in Fig.~2a.
Numbers in bold show the best performance for each metric and class, with a 2\% tolerance, excluding UpperBound.
Dice is shown as \%, and UVD as $\mu m^3$x$10^{2}$.
}
\centering
\scriptsize
\setlength{\extrarowheight}{.3ex}
\sisetup{detect-all=true}
\begin{tabular}{|l|
S[table-format=-2.2]
S[table-format=-2.2]|
S[table-format=-2.2]
S[table-format=-2.2]|
S[table-format=-2.2]
S[table-format=-2.2]|
S[table-format=-2.2]
S[table-format=-2.2]|
}
\hline
\multirow{2}{*}{ Method } & \multicolumn{2}{c|}{IRF} & \multicolumn{2}{c|}{PED} & \multicolumn{2}{c|}{SHRM} & \multicolumn{2}{c|}{SRF} \\
\cline{2-9}
& {~~Dice} & {~~UVD} & {~~Dice} & {~~UVD} & {~~Dice} & {~~UVD} & {~~Dice} & {~~UVD} \\
\hline
UpperBound[$D^\mathrm{t}$] & 69.33 & 10.44 & 65.23 & 7.09 & 60.03 & 7.82 & 60.93 & 9.32 \\
Baseline[$D^\mathrm{s}$] & 32.05 & 19.44 & 30.87 & 16.12 & 33.06 & 16.90 & 42.28 & 17.92 \\
\hdashline
SimCLR($P_\mathrm{augm}$,$C_\mathrm{pool}$)  & 61.79 & 12.84 & 47.88 & 11.42 & 41.44 & 16.33 & 43.21 & 12.84\\
SimSiam($P_\mathrm{augm}$,$C_\mathrm{pool}$) & 54.12 & 16.51 & 37.41 & 14.10 & 40.06 & 13.99 & 52.31 & 16.22  \\
\hdashline
SegCLR($P_\mathrm{augm}$,$C_\mathrm{pool}$) & 63.28 & 14.29 & 52.32 & 12.32 & \B 63.63 & \B  10.54 & 51.90 & 9.57  \\
SegSiam($P_\mathrm{augm}$,$C_\mathrm{pool}$) & 8.31 & 24.51 & 4.93 & 145.44 & 32.32 & 17.50 & 5.11 & 75.30 \\
SegCLR($P_\mathrm{slice}$,$C_\mathrm{pool}$) & 33.76 & 18.75 & 52.92 & 10.18 & 41.95 & 11.87 & 34.21 & 18.37 \\
SegCLR($P_\mathrm{comb}$,$C_\mathrm{pool}$) & \B 72.20 & \B 10.90 & \B 61.63 &\B  9.40 & 53.55 & 12.90 & 59.70 & 12.18 \\
SegCLR($P_\mathrm{comb}$,$C_\mathrm{ch}$) & 68.98 & 11.38 & 55.40 & 11.96 & \B 62.74 & 11.17 & \B 62.20 & \B 9.03 \\
\hline
\end{tabular}
\label{tab:results_cirrus}
\end{table}

\begin{table}[htb]
\caption{
Evaluation of models on source domain $D^\mathrm{s}$.
This table corresponds to the results in Fig.~2b.
Numbers in bold show the best performance for each metric and class, with a 2\% tolerance.
Dice is shown as \%, and UVD as $\mu m^3$x$10^{2}$.
}
\centering
\scriptsize
\setlength{\extrarowheight}{.3ex}    
\sisetup{detect-all=true}
\begin{tabular}{|l|
S[table-format=-2.2]
S[table-format=-2.2]|
S[table-format=-2.2]
S[table-format=-2.2]|
S[table-format=-2.2]
S[table-format=-2.2]|
S[table-format=-2.2]
S[table-format=-2.2]|
}
\hline
\multirow{2}{*}{ Method } & \multicolumn{2}{c|}{IRF} & \multicolumn{2}{c|}{PED} & \multicolumn{2}{c|}{SHRM} & \multicolumn{2}{c|}{SRF} \\
\cline{2-9}
& {~~Dice} & {~~UVD} & {~~Dice} & {~~UVD} & {~~Dice} & {~~UVD} & {~~Dice} & {~~UVD} \\
\hline
Baseline[$D^\mathrm{s}$] & 77.15 & 2.28 & 71.28 & 5.55 & 57.47 & 7.17 & \B 63.53 & \B 8.20 \\
\hdashline
SimCLR($P_\mathrm{augm}$,$C_\mathrm{pool}$) & 73.97 & 2.55 & 67.44 & 6.57 & 53.30 & 7.36 & 60.82 & 8.67  \\
SimSiam($P_\mathrm{augm}$,$C_\mathrm{pool}$) & \B 79.88 & \B 2.24 & \B 72.10 & 5.70 & 56.41 & 7.31 & 62.62 & 8.73  \\
\hdashline
SegCLR($P_\mathrm{augm}$,$C_\mathrm{pool}$) & 75.33 & 2.47 & 71.12 & 5.18 & 58.34 & 6.90 & 62.04 & 8.64  \\
SegSiam($P_\mathrm{augm}$,$C_\mathrm{pool}$) & 31.96 & 4.33 & 4.88 & 129.08 & 44.79 & 12.69 & 1.50 & 70.35  \\
SegCLR($P_\mathrm{slice}$,$C_\mathrm{pool}$) & 52.66 & 3.10 & 63.60 & 9.58 & 46.22 & 8.94 & 46.42 & 10.63 \\
SegCLR($P_\mathrm{comb}$,$C_\mathrm{pool}$) & \B 79.78 & 2.41 & \B 73.28 & \B 4.92 & 57.74 & 7.99 &\B 64.54 & 8.59 \\
SegCLR($P_\mathrm{comb}$,$C_\mathrm{ch}$) & \B 80.18 & 2.48 & 71.25 & 5.33 & \B 62.12 & \B 6.74 & \B 63.58 & \B 8.29  \\
\hline
\end{tabular}
    \label{tab:results_spectralis}
\end{table}

\begin{table}[htb]
    \caption{
    Balance between supervised and contrastive losses.
    Evaluation of the proposed
    SegCLR($P_\mathrm{comb}$,$C_\mathrm{ch}$) model with different values of the weighting parameter $\lambda$.
    Metrics here are calculated across all classes, relative to the same model with $\lambda$$=$$20$ used in all other experiments.
    Very low and high values (i.e., $\lambda=\{0.1,1,1000\}$) lead to substantially worse Dice and UVD metrics on both domains.
    Values closer to $\lambda$$=$$20$ (i.e., $\lambda=\{10,100\}$) have only a minor negative effect on the segmentation metrics.
    Dice is shown as \%, and UVD as $\mu m^3$x$10^{2}$.
}
    \centering
\setlength{\extrarowheight}{.3ex}  
\sisetup{detect-all=true}
\begin{tabular}{
|S[table-format=4.1]|
S[table-format=-2.2]
S[table-format=-2.2]|
S[table-format=-2.2]
S[table-format=-2.2]|
}
\hline
& \multicolumn{2}{c|}{$D^\mathrm{t}$} & \multicolumn{2}{c|}{$D^\mathrm{s}$} \\
\cline{2-5}
$\lambda$ & {~~Dice} & {~~UVD} & {~~Dice} & {~~UVD} \\
\hline
0.1 & -12.17 & 1.25 & -14.67 & 2.62 \\
1 & -14.84 & 2.11 & -15.20 & 1.28 \\
10 & -4.64 & 0.57 & -1.03 & 0.28 \\
100 & -1.41 & -0.03 & -0.01 & 0.26 \\
1000 & -8.65 & 2.01 & -2.31 & 0.31 \\
\hline
\end{tabular}
    \label{tab:lambda_ablation}
\end{table}
\end{document}